\documentclass[twocolumn]{article}
\usepackage[pdftex]{graphicx}
\usepackage{my_symbols}
\usepackage{url}			
\usepackage{color}			

\usepackage{amsthm} 
\newtheorem{proposition}{Lemma}

\usepackage{subfig}

\usepackage[normalem]{ulem}
\newcommand{\cout}[1]{}		

\begin{document}


	\title{Concept-drifting Data Streams are Time Series; \\ The Case for Continuous Adaptation}

\author{Jesse Read}
\date{}


\maketitle

\begin{abstract}
	Learning from data streams is an increasingly important topic in data mining, machine learning, and artificial intelligence in general. A major focus in the data stream literature is on designing methods that can deal with concept drift, a challenge where the generating distribution changes over time. A general assumption in most of this literature is that instances are independently distributed in the stream. In this work we show that, in the context of concept drift, this assumption is contradictory, and that  
	the presence of concept drift necessarily implies temporal dependence; and thus some form of time series. This has important implications on model design and deployment. We explore and highlight the these implications, and show that Hoeffding-tree based ensembles, which are very popular for learning in streams, are not naturally suited to learning \emph{within} drift; and can perform in this scenario only at significant computational cost of destructive adaptation. 
	On the other hand, we develop and parameterize gradient-descent methods and demonstrate how they can perform \emph{continuous} adaptation with no explicit drift-detection mechanism, 
	offering major advantages in terms of accuracy and efficiency. As a consequence of our theoretical discussion and empirical observations, we outline a number of recommendations for deploying methods in concept-drifting streams.
\end{abstract}


\section{Introduction}
\label{sec:intro}


Predictive modeling for data streams is becoming an increasingly-relevant task, in particular with the increasing advent of sensor networks and tasks in artificial intelligence, including robotics, reinforcement learning, system monitoring, anomaly detection, social network and media analysis. 

In a data stream, we assume that data arrives
\[
	(\x_t,y_t) \sim p_t(\dX,\dY)
\]
over time $t=1,\ldots,\infty$. A model observes test instance $\x_t$ and is required to make a prediction 
\[
	\yp_t = h_t(\x_t)
\]
\emph{at time} $t$. Hence the amount of computational time spent per instance must be less that the rate of arrival of new instances (i.e., the real clock time between time steps $t-1$ and $t$). A usual assumption is that true label $y_{t-1}$ becomes available at time $t$, thus allowing to update the model. 

This is in contrast to a standard batch setting, where a dataset $\{\x_t,y_t\}_{t=1}^N$ of fixed $N$ is observed prior to inducing the model. 
See \cite{IndresSurvey} for introduction and definitions. 

%


\subsection{Building Predictive Models from Data Streams}
\label{sec:learning_from_data_streams}

We wish to build a model that approximates, either directly or indirectly, the generating distribution. For example, a MAP estimate for classification
\[
	\yp_t = h_t(\x_t) = \argmax_{y} \papprox_t(\x_t,y)
\]
The incremental nature of data streams has encouraged a focus on fast and incremental, i.e., updateable, models; both in the classification and regression contexts. \cout{Although batch classifiers can sometimes be employed competitively (so-called `batch-incremental'; see, e.g., \cite{Kolter2007} -- and \cite{IDA2012} for an empirical comparison) the literature shows a clear preference for incremental models where the learning model is updated one instance at a time. }
Incremental decision trees such as the Hoeffding tree (HT, \cite{HT}) have had a huge impact on the data-streams literature and dominate recent output in this area. They are fast, incremental, easy to deploy, and offer powerful non-linear decision boundaries. Dozens of modifications have been made, including ensembles \cite{LeveragingBagging,EnsembleDriftSurvey,DataStreamEnsembleSurvey,AdaptiveRF}, and adaptive trees \cite{HAT}. 

High performance in data streams has also been obtained by $k$-nearest neighbors ($k$NN) methods, e.g., \cite{SAMkNN,MicroClusterKNN,kNNstream}. As a lazy method, there is no training time requirement other than simply storing examples to which -- as a distance-based approach -- it compares current instances, in order to make a prediction. The buffer of examples should be large enough to represent the current concept adequately, but not too large as to be prohibitively expensive to query.

Methods employing stochastic gradient descent (SGD)\footnote{i.e., incremental gradient descent; SGD usually implies drawing randomly from a dataset in i.i.d.\ fashion. Typically a stream is assumed to be i.i.d., and thus equivalent in that sense. We challenge the i.i.d.\ assumption later, but keep this terminology to be in line with that of the related literature} have been surprisingly underutilized in this area of the literature. Baseline linear approaches obtained relatively poor results, 
but can be competitive with appropriate non-linearities \cite{RTStreams2} and have been used within other methods, e.g., at the leaves of a tree \cite{SGDleaves}. In this work, we argue that the effectiveness of SGD methods on data-stream learning has been underappreciated and in fact offer great potential for future work in streams. 




\subsection{Dealing with Concept Drift}
\label{sec:drift}

Dealing with \textit{concept drift}, where the generating distribution $p_{t-1} \neq p_{t}$ in at least some part of the stream, is a major focus of the data-stream literature, since it means that the model current $h_t$ has become partially or fully invalid. Almost all papers on the topic propose some way to tackle its implications, e.g., \cite{HOFER2013377,TransferLearningForDrift,IndresSurvey,LeveragingBagging,EnsembleDriftSurvey,KatakisRecurring,Kolter2007}. A comprehensive survey to concept drift in streams is given in \cite{IndresSurvey}. 



The limited-sized buffer of $k$NN  methods imply a natural forgetting mechanism where old examples are purged, as dictated by available computational (memory and CPU) resources. \cout{A similar principle applies to ensembles of batch models (i.e., batch-incremental), where older or less-relevant models from the ensemble are dropped as new ones are created.} Any impact by concept drift is inherently temporary in these contexts. Of course, adaptation can be increased by flushing the buffer when drift is detected.

HTs can efficiently and incrementally build up a model over an immense number of instances without needing to prune or purge from a batch. However, a permanent change in concept (generating distribution) will permanently invalidate the current tree (or at least weaken the relevance of it -- depending on severity of drift) therefore dealing with drift becomes essential. The usual approach is to deploy a mechanism to detect the change, and reset a tree or part of a tree when this happens, so that more relevant parts may be grown on the new concept. Common detection mechanisms include ADWIN \cite{ADWIN}, CUMSUM \cite{egCUMSUM}, Page Hinkley \cite{HanenPageHinkley}, and various geometric moving average and statistical tests \cite{Gama2004}. For example, the Hoeffding Adaptive Tree (HAT, \cite{HAT}) uses an off-the-shelf ADWIN detector at each node of a tree, and cuts a branch at a node where changes in registered. It is expected that the branch will be regrown on new data and thus represent the new concept. Tree approaches are almost universally employed in ensembles to mitigate potential fallout from mis-detections. 



In this paper we argue strongly for the potential importance of a third option -- of continuous adaptation, where knowledge (e.g., a set of parameters determining a decision boundary) is transferred as best as possible to a newer/updated concept rather than discarded or reset as in currently the case with the popular \emph{detect and reset} approach. This possibility can be enacted with SGD. SGD is intimately known and widely used across swathes of the machine learning literature, however, we note that it is markedly absent from the bulk of the data-stream methods, and often only compared to only as a baseline in experiments. In this paper we argue that it has been discarded prematurely and underappreciated. We analyse and parameterize it specifically in reflection to performance in concept-drifting data-streams, and show it to be very competitive with other approaches in the literature (results in \Sec{sec:discussion}). 

To summarize the main mechanisms to deal with concept drift:
\begin{enumerate}
	\item Forgetting mechanism\\ (e.g., $k$NN, and batch-incremental ensembles)
	\item Detect and reset mechanisms\\ 
		(e.g., HATs and HT-ensembles with ADWIN)
	\item Continuous adaptation\\ (e.g., SGD-based models incl.\ neural networks)
\end{enumerate}

\subsection{Organization and contributions}



In spite of the enormous popularity of ensembled trees and distance-based ($k$NN) approaches, we will show that continuous adaptation can be more suited to concept-drifting data streams. We do this by breaking with a common assumption. 

Namely, existing work in data streams is mostly based on the assumption of i.i.d.\ data within a particular concept; therefore seeking as an objective to detect a change in concept, so that off-the-shelf i.i.d.\ models can be (re)-deployed. A model belonging to a previous concept is seen as invalid. This leads to the detect-and-reset approach to dealing with concept drift mentioned above. In this work, on the contrary, we show that drift inherently implies temporal dependence; that all concept-drifting streams are in some way a time series, and can be treated as such. We propose to treat the concept as a temporal sequence, to enable continuous adaptation as an effective alternative to detect-and-reset approaches. For this purpose we derive gradient descent approaches; and we show scenarios where they compare very favorably in comparison with more popular tree and distance-based methods. 
%
%
The contributions of this work can be summarized as follows: 
\begin{itemize}
	\item We show that concept-drifting data streams can be considered as time series problems (\Sec{sec:time_series})
	\item Following a fresh analysis on concept drift (\Sec{sec:revision}), we conduct a bias-variance analysis of learning under drift, and derive gradient descent approaches for concept-drifting data streams in the framework of continuous adaptation (\Sec{sec:novel})
	\item We give an analytical and empirical investigation (\Sec{sec:experiments}) which highlights properties of our suggested approach; and from the results  (displayed and discussed in \Sec{sec:discussion}) we outline the implications and make a number of recommendations before making concluding remarks (\Sec{sec:conclusion}).
\end{itemize}

\section{Concept-drifting 
Streams are Time Series}
\label{sec:time_series}

Concept-drifting data streams have been widely treated under the assumption of independent samples (see, e.g., \cite{Gama2004,LeveragingBagging,IndresSurvey,Persistent} and references therein). In this section, we argue that if a data stream involves concept drift, then the independence assumption is violated. 

\begin{figure}
	\centering
	\subfloat[][]{
		\includegraphics[scale=0.9]{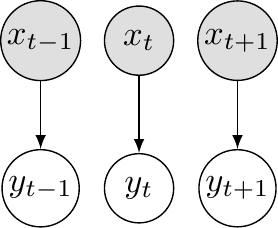}
		\label{fig:main_figure.a}
	}
	\quad
	\subfloat[][]{
		\includegraphics[scale=0.9]{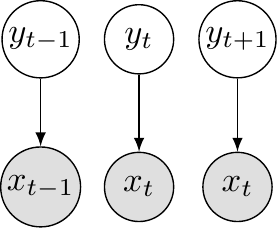}
		\label{fig:main_figure.a_bis}
	}
	\caption{\label{fig:pre_main_figure}The `standard' discriminative \fref{fig:main_figure.a} and generative \fref{fig:main_figure.a_bis} models for prediction in data streams; as directed probabilistic graphical models.}
\end{figure}

If data points are drawn independently, we should be able to write
\begin{align}
	p_t(y_t,x_t,x_{t-1}) &= p_t(y_t|x_t)p_t(x_{t-1}) \label{eq:ind}\\ 
	p_t(y_t|x_t,x_{t-1}) &= p_t(y_t|x_t) \notag 
\end{align}
see \Fig{fig:main_figure.a} for illustration, where the lack of an edge between time-steps indicates the independence. 
%
%
\cout{
Indeed, we can use the joint distributions by \Fig{fig:pre_main_figure} to derive probabilistic methods, such as the naive Bayes (a typical data-streams baseline classifier): 
\begin{equation}
	\label{eq:map}
	\yp_t = h(\x_t) = \argmax_{y_t} P(y_t,\x_t)
\end{equation}
}

The subscript of $p_t$ reminds us of the possibility of concept drift (in which case, $p_t \neq p_{t-1}$). 


\begin{proposition}
	\textit{A data stream that exhibits a concept drift also exhibits temporal dependence.} 
\end{proposition}


\begin{proof}
	Let $p_t(y_t|x_t) = p(y_t|x_t,C_t)$ where $C_t \in \{0,1\}$ denotes the concept at time-step $t$, 
	and let the drift occur at point $0 < \tau < \infty$. Thus $C_t = 0$ for $t < \tau$, and $C_t = 1$ for $t \geq \tau$. Under independence,
	\[
		P(C_{t}) = P(C_{t} | C_{t-1})
	\]
	However, it is obvious that 
	\[
		P(C_{t}=0) \neq P(C_{t}=0 | C_{t-1}=1),
	\]
	namely, after the drift we no longer expect instances from the first concept. 

	We can use the joint distribution, \Eq{eq:ind}, to check for independence, for any particular time step $t$, 
	marginalizing out $C$ which is not observed:
	
\begin{align}
	p(y_t|x_t,x_{t-1}) &\propto p(y_t,x_t,x_{t-1})  \nonumber \\
	                   &= \sum_{c_t,c_{t-1}} p(y_t,x_t,c_t, c_{t-1}, x_{t-1})  \nonumber \\
		   &= p(y_t|x_t)\underbrace{ \sum_{j=1}^{t}  \sum_{c_j,c_{j-1}}p(x_{j}|c_{j})p(c_j|c_{j-1})}_{p(x_t|x_{t-1})} \nonumber \\
			&= p(y_t|x_t)p(x_t|x_{t-1}) \label{eq:deriv} 
\end{align}
	which \emph{does not} equal \Eq{eq:ind}, thereby indicating temporal dependence (via the presence of $t-1$). 
\end{proof}

\begin{figure}
	\centering
	\subfloat[][]{
		\label{fig:main_figure.b}
		\includegraphics[width=0.42\columnwidth]{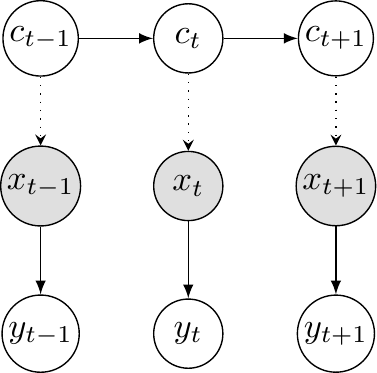}
	}
	\quad
	\quad
	\subfloat[][]{
		\includegraphics[width=0.42\columnwidth]{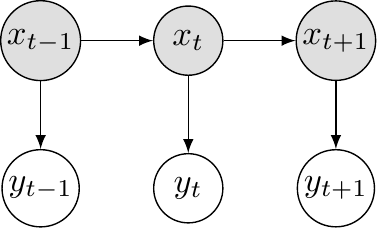}
		\label{fig:main_figure.c}
	}
	\caption{\label{fig:pre_main_figure2}A discriminative model \fref{fig:main_figure.b} for prediction in data streams with concept drift (dashed lines simply indicate that these connections are not usually explicitly addressed). When drift is taken into account and marginalized out \fref{fig:main_figure.c}, inputs become connected.
	}
\end{figure}

This can be visualized as a probabilistic graphical model in \Fig{fig:main_figure.b} (observed variables are shaded); and 
in \Fig{fig:main_figure.c} with concept drift marginalized out. 

We have shown above that data streams with concept drift exhibit temporal dependence, which essentially means that all such streams can be seen as a time series. 

We have done this analysis in the most extreme case of a switch in concepts over a single time step (a sudden change in concept). One might argue that as $t \rightarrow \infty$, the importance of the dependence resulting from this one-time drift becomes negligible, since \emph{after} drift, $P(C_t=1)=P(C_{t-1})=1$ and could thus be considered a constant essentially rendering independence \emph{within} each concept. However we do not observe $\tau$; we cannot know exactly when the drift will occur or if it has ocurred. As a result, an instantaneous drift between two time steps can manifest itself as temporal dependence in the error signal over many instances. It is surprising that this is not explicit across the literature, since it is indeed implicit in most change-detection algorithms, 
in the sense that they measure the change in the error \emph{signal} of predictive models.







\begin{figure}
	\centering
	\subfloat[][]{
		\includegraphics[width=0.32\columnwidth]{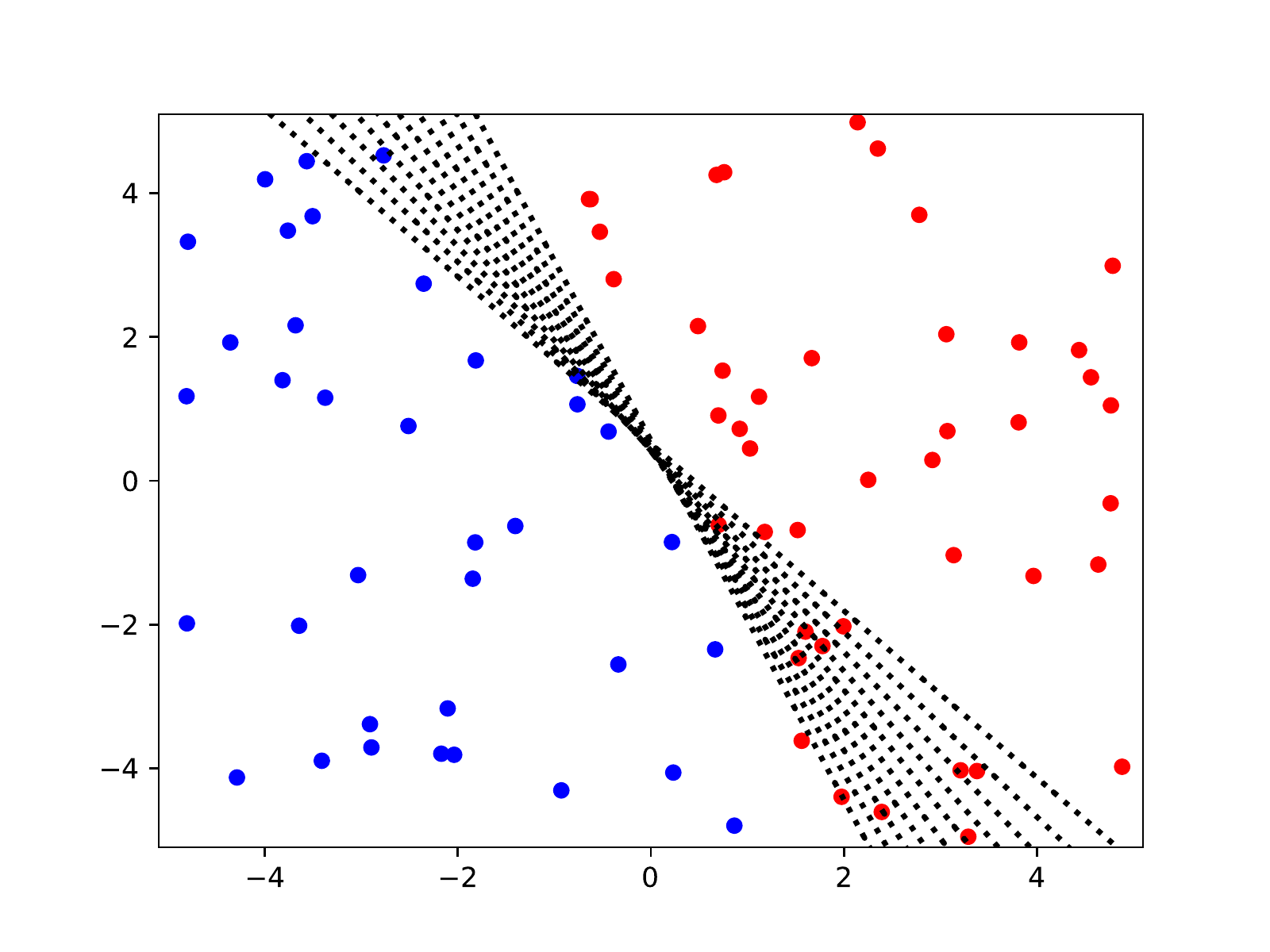}
		\label{fig:drift_series.a}
	}
	\subfloat[][]{
		\includegraphics[width=0.32\columnwidth]{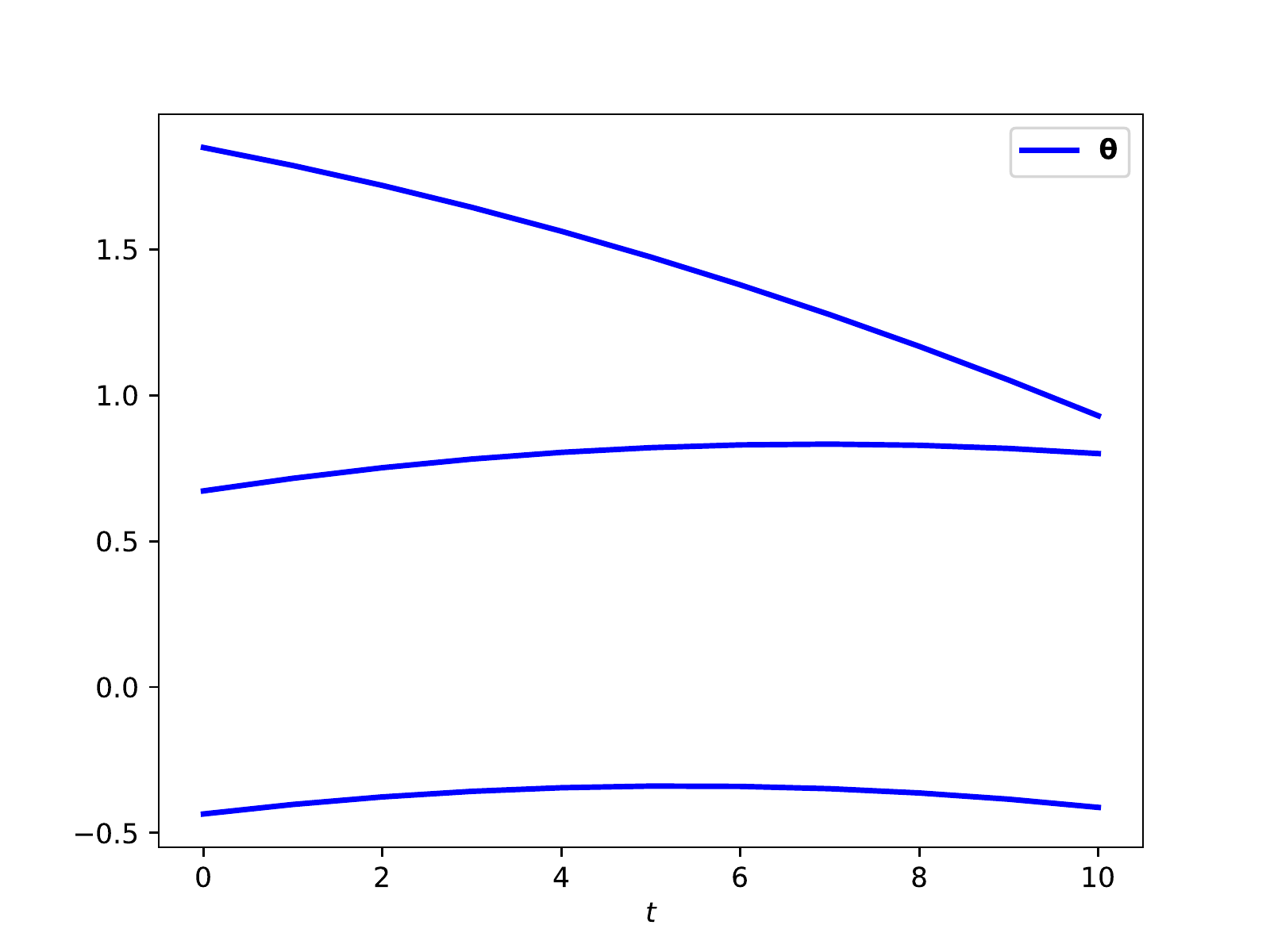}
		\label{fig:drift_series.b}
	}
	\subfloat[][]{
		\includegraphics[width=0.32\columnwidth]{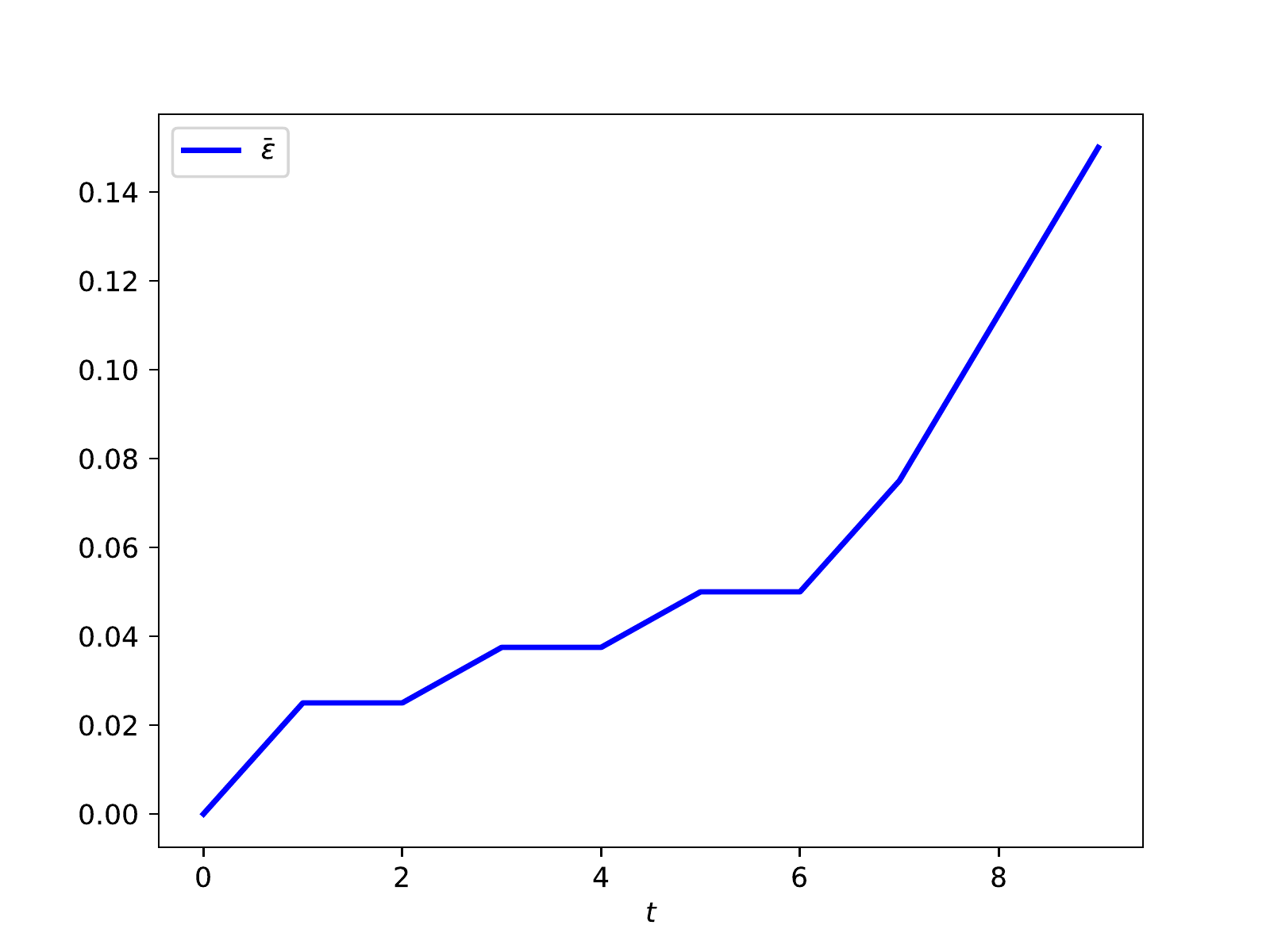}
		\label{fig:drift_series.c}
	}
	\caption{\label{fig:drift_series}Concept drift (incremental drift): decision boundary on $y_t = \sigma(\theta_t^\top\x_t)$ shown for $t=\tau,\tau+1,\ldots,\tau+10$ \fref{fig:drift_series.a}; corresponding value of the coefficients over these time-steps \fref{fig:drift_series.b} -- forming a time series with obvious temporal dependence: $\theta_\tau,\ldots,\theta_{\tau+10}$; and the error rate over time -- also a clear time series \fref{fig:drift_series.c}. }
\end{figure}

The relationship between the predictive model and the error is clearly seen in the relation
\[
	E_t =E(h_t(\x_t), y_t)
\]
where $E$ is the error function (e.g., mean-squared or classification error, depending on the problem). Clearly if $h_t$ is poorly adapted to deal with a concept drift, this will appear in increasing $E_t$ (i.e., a time series). It is illustrated (for incremental drift\footnote{We will review this type of drift in \Sec{sec:revision}}) in \Fig{fig:drift_series}. Rather than monitoring $\{E_t\}$ for drift so as to reset $h_t$, in this paper, we look at adapting $h_t$ directly. 




Having argued that concept drifting streams are time series -- should we just apply off-the-shelf time series models? Expanding on some differences mentioned in \cite{Persistent}, we can point out that
\begin{enumerate}
	\item Only data streams exhibiting \emph{concept drift} are guaranteed to have time series elements, and only in consideration specifically of these parts of the stream 
	\item Time dependence in data streams it is seen as a problem (something to deal with), rather than as part of the solution (something to explicitly model) 
	\item In data streams the \emph{final} estimation of each $\yp_t$ is required at time $t$, and thus retrospective/forward-backward (smoothing) inference (as is typical in state-space models) is not applicable
	\item A common assumption in data streams is that the ground truth is available at time $t-1$, providing a stream of training examples $(\x_{t-1},y_{t-1})$ whereas time series models are typically built offline before being deployed live \label{i:3}
	\item Data streams are assumed to be of infinite length (therefore, also training data, on account of item \ref{i:3})
\end{enumerate}
Some of these assumptions are broken on a paper-by-paper level. For example, changes to point \ref{i:3} have been addressed in, e.g., \cite{AdaptiveRF,HOFER2013377}. 

The most closely related time series task to prediction\footnote{Of the current time step} in streams is that known as \emph{filtering}. Actually, \Eq{eq:deriv} is a starting point for state space models such as hidden Markov models, Kalman and paricle filters (see \cite{Barber} for a thorough review) for which \emph{filtering} is a main task, although these models are not usually expected to be updated at each timestep as in the streaming case (a possibility due to point \ref{i:3} above). 

Although we cannot always apply state-space models directly to data streams, we nevertheless remark again that temporal dependence plays a non-negligible role, and can be leveraged as an advantage. In particular, in reflection of this section, we next make a revision of concept drift -- in \Sec{sec:revision} -- which then allows us to employ efficient and effective methods (\Sec{sec:novel}), allowing us to draw conclusions that have important repercussions in data streams mining.

\section{Types of Concept Drift: A Fresh Analysis}
\label{sec:revision}

Sudden, complete, and immediate drift is widely considered in the literature (and for this reason we worked with it in the previous section), yet we could also argue that gradual, or incremental drift fit better a dictionary definition which implies a \emph{movement} or tendency\footnote{Cambridge dictionary offers, among others: ``\textit{a slow movement}''} and are more widespread in practice. A complete change in concept simply means we have changed problem domains.  
The idea of a movement inherently includes the implication of dependence (e.g., across time and space) -- and dependence (among instances) implies naturally a time \emph{series} -- as we have elaborated in the previous section.

Let us review and revise concept drift, based on the types of drift identified in \cite{IndresSurvey}, which relist for convenience:
\begin{enumerate}
	\item Sudden/abrupt \label{l:1}
	\item Incremental \label{l:2}
	\item Gradual \label{l:3}
\end{enumerate}
additionally noting the possibility of reoccurring drift which may involve any of these types and, noting also the related task of dealing with outliers, which are \emph{not} concept drift. 


In the following, we denote $\theta_t \in \Theta$ as the \emph{true} (unknown) parameters defining the current concept at time $t$, i.e., $p_{\theta_t}$. This allows for a smooth \emph{drift} across concept space $\Theta$ (for example, a set of coefficients defining a hyperplane; see \Fig{fig:drift_series}) but also allows for a qualitative view of categorical concepts $C_t \in \{1,\ldots,K\}$; such that $\theta^{c}$ represents the parameters of the $c$-th concept; i.e., we would speak of drift between concepts $\theta^{{1}}$ and  $\theta^{{2}}$.

\subsection{Abrupt change}

If the concept changes abruptly, in either a partial or complete manner, we may denote
\begin{equation}
	\label{eq:1}
	\theta_t = \CaseCond{\theta^{c_1}}{t < \tau}{\theta^{c_2}}{t \geq \tau}
\end{equation}
for some time index $\tau$ where the `drift'\footnote{As noted above, we would prefer the term \emph{shift} for this particular case. Nevertheless, we inherent this terminology from the literature for the sake of consistency} occurs. The drift may be total ($\theta^1$ and $\theta^2$ are drawn independently from $\Theta$) or partial (only a local change, to some part\footnote{Recalling that $\theta$ is likely to be multi-dimensional} of $\theta$.


\subsection{Incremental drift}
\label{sec:d_2}




Incremental drift denotes a change over time. It can be denoted as an additive operation, where the current concept can be written as
\begin{equation}
	\label{eq:2}
	\theta_t = \theta_{t-1} +  \Delta_t\theta
\end{equation}
i.e., an increment of $\Delta_t\theta$. 
We generally assume that drift is active in range $\tau_1 \leq t \leq \tau_2$ (and that $\Delta \theta = 0$ outside of this range); where concept $c_1$ before, $c_2$ after, and a blended mixture inbetween). 

\subsection{Gradual drift}
\label{sec:d_3}

In gradual drift, drift also occurs over time, but in a stochastic way. We may write
\begin{equation}
	\label{eq:3}
	\theta_t = \theta^{c_t} \quad\text{ where }\quad {c_t} \sim \dB(\alpha_{t}) 
\end{equation}
where $c_t=1$ with probability $\alpha_t$, and $c_t=2$ otherwise; $\dB$ being a Bernoulli distribution. Note that $\alpha_{t}$ is itself is an incremental drift (see \Eq{eq:2}) between values $0$ and $1$. The stream thus increasingly-often generates examples from the new concept $c_2$; where drift $\alpha_{t < \tau_1} = 0$ and $\alpha_{t > \tau_2} = 1$ outside of the drift range $\tau_1,\ldots,\tau_2$. 

Note that neither incremental nor gradual drift need be smooth or monotonically increasing, although that is a common simplification. A sigmoid function is often used in the literature; as in many of the stream generators of the MOA framework \cite{MOA}. 

\subsection{Re-occurring drift}
\label{sec:reoccurring_drift}

Re-occurring drift may be any of the above cases (sudden, gradual, or incremental) where a concept may repeat at different parts of the stream. 
It is very much related to the idea of state-space models such as hidden Markov models, and switching models (both are reviewed in \cite{Barber}). 
We remark that there is no technical difference between modelling states, and tracking concepts. Usually we can distinguish a state as something that we \emph{want to model} (e.g., a weather system), and a concept drift as something we wish to adapt to or take into account (e.g., change/degradation of in the monitoring sensors, or climate change).

%


\section{Learning under Drift: Theoretical Insights}
\label{sec:novel}

In this section we investigate an approach to adapt to drift continuously as part of the learning process, rather than reactively (i.e., the detect-and-reset approach) using explicit drift detection mechanisms, as has previously been the main approach (see \Sec{sec:intro}). 


It is well known that prediction error of supervised learners breaks down into \textit{variance}, \textit{bias}, and \textit{irreducible error} (see, e.g., \cite{Barber,DuSwamy}). Let $f : \dX \rightarrow \R$ represent the true underlying (unknown) model parametrized by $\theta_c$, which produces observations $y_t = f(\x_t; \theta_c) + \epsilon_t$, where $\epsilon_t \sim \N(0,\sigma^2)$. The expected mean squared error (MSE) over the data, with some estimated model $h$, can be expressed as
	\begin{align}
		\Exp_{\x_t}[\textsf{MSE}(h,y|\x_t)] & = \Exp_{\x_t} \Exp_{y_t, \epsilon_t}[ (y_t - h)^2 | \x_t]\nonumber \\
			&= \sigma^2 + \Vxp[h] + \Exp[f - h]^2 \label{eq:bvt_normal}
	\end{align}
i.e., irreducible noise, variance, and bias$^2$ terms, respectively. 

The result in \Eq{eq:bvt_normal} hinges on the assumption that 
		\(
			\Exp[y] = \Exp[f + \epsilon] = f 
		\) 
		due to the fact that $\Exp[\epsilon_t] = 0$. 

However, recall that in the case of concept drifting data streams, $f$ is not constant, but inherits randomness from random variable $C_t$ (see Sections~\ref{sec:time_series} and \ref{sec:revision}). We can get around this problem by 
taking the expectation of $\textsf{MSE}$ within each concept; wrt the point of a single change (which we denote $\tau$), then for time $t \geq \tau$),
\[
	y_t = \underbrace{\left[f_{t<\tau} + \Delta(f_{t<\tau},f_{t\geq\tau})\right]}_{f_{t \geq \tau}}(\x_t) + \epsilon_t
\]
where $\Delta(f_{t<\tau},f_{t\geq\tau})$ represents the change in concept. Regarding \Eq{eq:bvt_normal}: At $t=\tau$ the 
third term (bias) is now essentially the difference between the current true concept, and an estimated old concept. 
In other words (in terms of parameters), 
\begin{equation}
	\label{eq:tracking}
	(\theta_t - \thest_t)^2
\end{equation}
and clearly the obvious goal (in terms of reducing bias), is moving $\thest_t$ (an estimate of the previous concept) towards $\theta_2$ (the true \emph{current} concept), and over a concept drifting stream of multiple drifts of different types: to model the journey of $\theta$.

In the data streams literature, ensemble models with drift-detection strategies have blossomed (see, \cite{DataStreamEnsembleSurvey,EnsembleDriftSurvey} for surveys). 
We can now describe theoretical insights to this popularity in this particular area\footnote{Aside from the popularity and effectiveness of ensembles in supervised learning in general}: By resetting a model when drift is detected, it is possible to reduce the bias implied by the drift in those models. However, this may increase the variance component of the error (since variance can be higher on smaller datasets \cite{SmallDataVariance}). Ensembles are precisely renown for reducing the variance component of the error (see., e.g., \cite{DuSwamy} and references therein) and thus desirable to counteract it. 

This analysis also leads us once more to the downside of this approach: Under long and intensive drift, a vicious circle develops; increased efforts to detect drift lead to more frequent detections and thus frequent resetting of models (implicitly, to reduce bias), which encourages the deployment of ever-larger ensembles (to reduce the variance caused precisely by resetting models). As seen in the literature, ensemble sizes continue to grow; and our experiment section confirm that such implementations can require significant computational time. 

Methods with an explicit forgetting mechanism (e.g., $k$NN, batch-incremental methods -- also popular in the literature) will automatically establish `normal' bias as $(t - \tau)$ becomes as large as the maximum number of instances stored in memory. However, this can be a long time if that size is large; and not a solution when the drift is sustained over a long time or occurs regularly.

Finally, we remark again that detectors will fire when the error signal has \emph{already} shown a significant change, 
by which time many (possibly very biased) predictions may have been made. 

In the following section we discuss how to avoid this trade-off. Namely, we propose strategies of continuous adaptation which do not detect drift, but \emph{track} drift through time. 



\section{Continuous Adaptation under Concept Drift}
\label{sec:proposal}





Drift detection methods usually monitor the error signal retrospectively for change (is $\epsilon_t$ statistically different from $\epsilon_{t-1}, \epsilon_{t-2}, \ldots$; a warning can be raised). However, since -- as argued above -- concept drift can be seen as a time series we can attempt to forecast and track the drift, and adapt continuously.

	In this sense, solving the concept drift problem is identical to solving the forecasting problem of predicting $\theta_t$ which defines the true unknown concept (see \Eq{eq:tracking}). 
Using all the stream we have seen up to timestep $t$, we could write 
\[
	\thest_{t} = g(\x_{1},\ldots,\x_{t}, y_{1},\ldots,y_{t-1})
\]
which at first glance appears unusable because it is a function over an increasingly large window of data. However, using a recursion on $\thest_{t-1}$, for minimizing least mean squares, there is a closed-form solution: \emph{recursive least squares} (RLS), which uses recursion to approximate 
\[
	\thest_t = \thest_{t-1} + \inv{\mR}_t \x_t (y_t - \x_t^\top \thest_{t-1})
\]
where $\inv{\mR}_t = \inv{\mR}_{t-1} - \inv{\mR}_{t-1}\x_t\inv{(1 + \x_t^\top\inv{\mR}_{t-1}\x_t)} \x^\top_{t}\inv{\mR}_{t-1}$. 

RLS is a well known adaptive filter, which can be easily extended to \emph{forgetting}-RLS and Kalman filters (see, e.g., \cite{DuSwamy}). 

Noting that $\x_t (y_t - \x_t^\top \thest_{t-1}) = \nabla E_{\thest_t}$ and replacing $\inv{\mR}_t$ with learning rate $\lambda$ we derive stochastic gradient descent (SGD), as follows. 


For $t=1,\ldots,\infty$:
	\begin{align*}
		\yp_t            & = h(x_t;\thest_t) \\
		\Delta \thest_t     &= \lambda \nabla E_{\thest_t}(\yp_{t},y_{t}) \\
		\thest_{t+1}     & \gets \thest_{t} + \lambda \nabla  E_{\thest} = \thest_t + \Delta \thest_t
	\end{align*}
where the last line is essentially forecasting the concept for the following timestep (note the connection to, e.g., \Eq{eq:2}). Note the data-stream assumption that at time-point $t+1$ we have already observed the true value $y_t$. 

This may be viewed as a trivial result, but it has important implications regarding much of the data-stream research and practice, which traditionally relies predominantly on ensembles of decision trees or $k$-nearest neighbor-based methods; see \Sec{sec:learning_from_data_streams}. We have argued that it has been underappreciated. Unlike Hoeffding tree methods, a gradient-descent based approach can learn from a concept-drifting stream without explicit concept drift detection. Unlike $k$NN, time complexity is much more favourable.

We remark that for SGD to perform robustly over the length of a stream, we have to ensure certain conditions. In particular, to \emph{not} decay the learning rate $\lambda$ towards zero over time. In batch scenarios wish wish to converge to a \emph{fixed} point and such learning rate scheduling is common and effective practice. However, under as stream this would cause SGD to react more and more slowly to concept drift until eventually becoming stuck in one concept. 

\begin{figure*}
	\centering
	\subfloat[][]{
		\label{f:a}
		\includegraphics[width=0.30\textwidth]{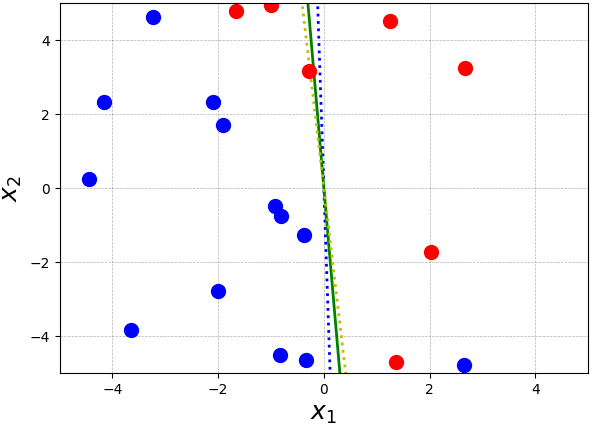}
	}
	\subfloat[][]{
		\label{f:b}
		\includegraphics[width=0.30\textwidth]{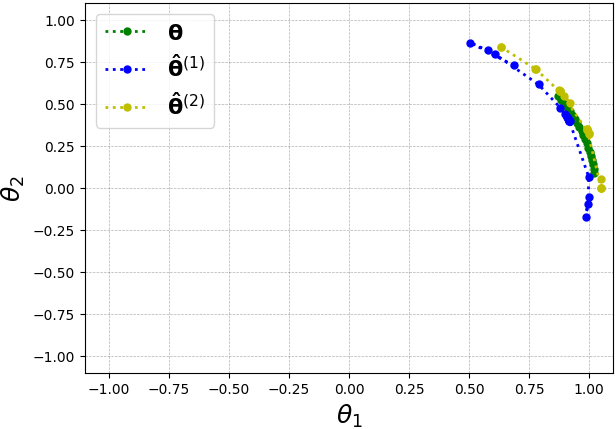}
	}
	\subfloat[][]{
		\label{f:c}
		\includegraphics[width=0.30\textwidth]{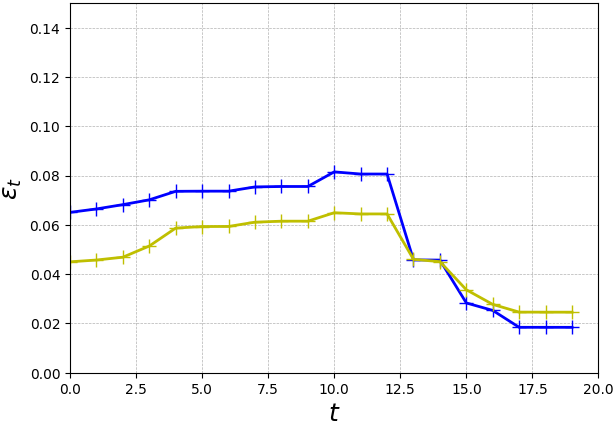}
	}
	\caption{\label{fig:missing} Data points from a stream exhibiting concept drift \fref{f:a}; where the true concept is represented as a decision boundary $\theta^\top\x = 0$ (
	shown in green) which is rotating clockwise; and estimates $\thest_t$ (in blue and yellow) are following it. Both are SGD methods with learning rate $\lambda=0.5$, and the blue has an additional momentum term $\beta = 0.5$. We see the path through weight/concept space \fref{f:b}, i.e., the \emph{concept drift}. Rather than a reactive approach monitoring the error rate retrospectively, $\thest_{t+1} \approx \theta_{t+1}$ can be tracked and pre-empted. 
Drift is constant across this time window, yet error rate $\epsilon_t|t=1,2,\ldots$ recedes \fref{f:c} as $\thest$ converges on the true moving concept. We emphasise that predictive performance improves \emph{during} the drift.}
\end{figure*}

An illustration of how SGD performs in a constant-drift setting is given in \Fig{fig:missing} on a synthetic data stream (which is detailed later in \Sec{sec:experiments}). We clearly see how no drift detection or model reset is necessary, and a concept can be smoothly tracked in its drift through concept space.


%
%
%
%

Recall that under gradual drift, it is the $\alpha_t$ (see \Eq{eq:3}) rather than $\theta_t$ which forms a time series. A detailed treatment is left for future work.

Since \emph{abrupt drift} cannot necessarily be forecast in advanced, one might argue that traditional drift detectors are best suited to this case. We remark that this argument can be clearly accepted only under the condition of a complete change in concept; where the two tasks are not at all related; a scenario unlikely to be the case in practice.  
If the drift is partial (i.e., the two concepts are partially related), then we wish to transfer part of the old concept (i.e., \emph{not} discard it when drift is detected). We note that SGD, in this sense, performs a kind of transfer learning; namely continuous transfer learning. The literature on transfer learning (see, e.g., \cite{TransferLearning}) indicates that we are thus likely to learn the new concept much faster.

\section{Experiments}
\label{sec:experiments}

\cout{In this section we design and carry out a number of experiments which will enrich the discussion following; namely to examine the strategies we outlined, in regards to concept drift and different types of synthetic and real-world stream settings. 
}



We carried out a series of experiments to enrich the discussion and support the arguments made in this work.  
All methods are implemented in Python and evaluated in the \textsc{Scikit-MultiFlow} framework \cite{MultiFlow} under prequential evaluation (testing then training with each instance). Experiments are carried out on a desktop machine, 2.60GHz, 16GB RAM.  


First, we generated synthetic data using a weight matrix $\theta_{c} \sim \N(\vec{0},\mI)$ to represent the $c$-th concept. We introduced drift using the equations \Eq{eq:1}, \Eq{eq:2}, \Eq{eq:3}, under parameters in \Tab{tab:synth}. 
\cout{We generate synthetic data of the dimensions detailed in \Tab{tab:synth}: a stream of $T$ instances, with concept drift from time $t=\tau_1,\ldots,\tau_2$, and an initial period from time $t=1,\ldots,\tau_0$ where accuracy is not yet recorded. \count{This pre-training is sound in terms of evaluation, since accuracy will be quite unstable at the beginning as methods are still `burning in' and realistic because in real-world applications some sample of data is typically available prior to launching a live system. The true concept $\theta_t$ is represented by a weight matrix, initially $\theta_1 \sim \N(\vec{0},\mI)$.} We then draw $\x_t \sim \N(0,1) | t = 1,\ldots$, and set corresponding label $y_t = \tup{\theta_t^\top\x_t \geq 0}$ (where $\tup{\cdot} = 1$ if the condition holds and $0$ otherwise). For incremental drift, we apply (in reference to \Eq{eq:2}),}
For incremental drift, 
\[
	\theta_t = A_{0.01}^\top\theta_{t-1} = \theta_{t-1} + \Delta_t\theta
\]
where $A_{0.01}$ is a rotational matrix (of angle $0.01$ in radians); for gradual drift, 
\[
	\alpha_t = p(c_t=1) = \frac{1}{\tau_2 - \tau_1} (t - \tau_1). 
\]
and, for sudden drift a new $\theta_t \sim \Theta$ is simply re-sampled after timestep $\tau_1$.

\begin{table}
	\centering
	\caption{\label{tab:synth}Parameters for synthetic data, except: $\tau_2=\tau_1+1$ in the case of sudden drift, and clearly $\tau_1,\tau_2$ are not used at all if no drift. In all experiments, accuracy is recorded over instances $\tau_0,\ldots,T$.}
		\begin{tabular}{lll}
			\hline
			Sym.         & $t$          & Description \\
			\hline
				$\tau_0$ & $T/10$       & pre-training ends \\
				$\tau_1$ & 5K           & start of drift\\
				$\tau_2$ & 6K           & end of drift (gradual, incr.)\\
				$\tau_2$ & $\tau_1 + 1$ & end of drift (sudden)\\
				$T$      & 10K          & length of stream\\
			\hline
		\end{tabular}
\end{table}

\cout{This synthetic data is designed to aid the interpretation of results in a relatively simplistic manner. One could argue that it favours SGD-based learners since it is based on a rotating decision boundary. Therefore, to test in a more competitive environment, we also generate more complex synthetic data of a different nature, namely by the \textit{random tree generator} (RTG, generator available in MOA \cite{MOA}; based on \cite{DomingosH00}). The RTG builds a random tree, assigning a random class to each leaf, then assigns random values to attributes which determine the classes via the tree structure. Clearly, such data should favour heavily decision tree learners like Hoeffding-tree based methods.}



We additionally look at two common benchmark datasets from the data-streams literature involving real-world data: the \textsf{Electricity} and \textsf{CoverType} datasets. \textsf{Electricity} contains 45,312 instances, with has $6$ attributes describing an electricity market (the goal is to predict the demand). \textsf{CoverType} contains 581,012 instances of 54 attributes, as input to predict one of seven classes representing forest cover type. See, e.g., \cite{Gama2004,IDA2012,HAT,AdaptiveRF} for details\footnote{The data is available at \url{https://moa.cms.waikato.ac.nz/datasets/}}.

\cout{
\subsection{Experimental Procedure and Results}
\label{sec:proc}
}

We employed the three main approaches discussed in this work (listed in \Tab{tab:methods}); both a vanilla configuration (`stardard' $k$NN, SGD, HT) and an `advanced' configuration of the same methodologies. For HT and $k$NN we used state-of-the-art adaptations. For SGD, we simply employed basis expansion to accommodate non-linear decision boundaries. 


\begin{table}
	\centering
	\caption{\label{tab:methods}Methods and their parameterization, both for a `standard' implementation (Vanilla Configuration) and high-performance (Advanced Configuration). Values for any parameters not shown can be found as the default parameters in \textsc{ScikitMultiflow} (v0.1.0 used here).} 
	\begin{tabular}{|lp{0.70\columnwidth}|}
		\hline
			$k$NN    & $k=10$, buffer size 100    \\                                         
			SGD      & $L_2$ regularization; $\lambda=0.01$, hinge loss  \\                  
			HT       & $10^{-7}$ split confidence, 0.05 tie threshold, naive Bayes at leaves \\ 
		\hline
		SAM$k$NN & Self-adjusting memory $k$NN \cite{SAMkNN} \\
		PBF-SGD& SGD with deg.\ $3$ polynomial basis expansion, e.g., \cite{DuSwamy} \\
		RF-HT & Adaptive Random Forest \cite{AdaptiveRF}: an ensemble of $100$ HTs, ADWIN drift detection, $\lambda=6$, $n_{min}=50$    \\
		\hline
	\end{tabular}
\end{table}

\begin{figure*}[h]
	\centering
	\subfloat[][Stationary concept]{
		\label{f:synth.a}
		\includegraphics[width=0.44\textwidth]{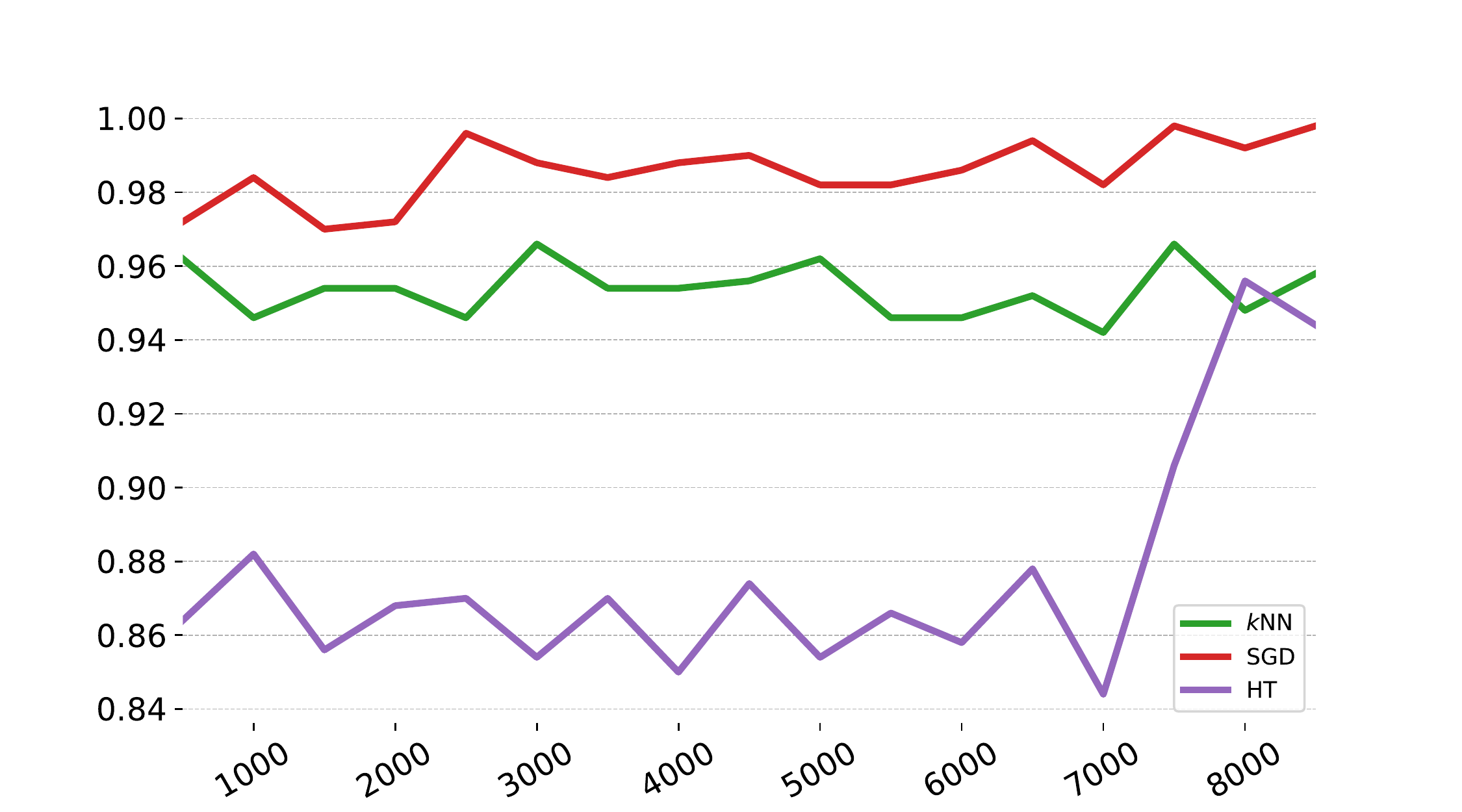}
	}
	\subfloat[][Sudden drift]{
		\label{f:synth.b}
		\includegraphics[width=0.44\textwidth]{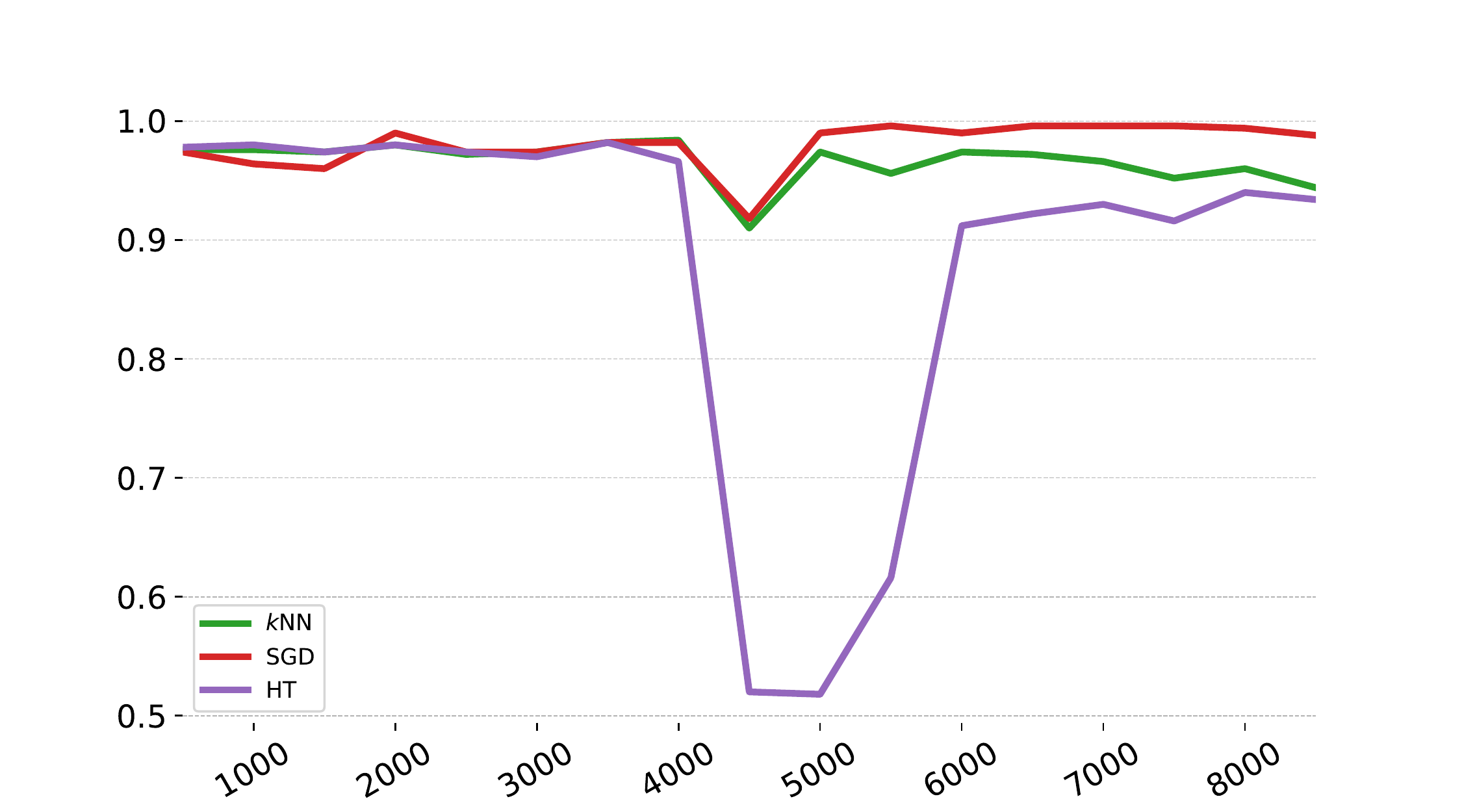}
	}\\
	\subfloat[][Incremental drift]{
		\label{f:synth.c}
		\includegraphics[width=0.44\textwidth]{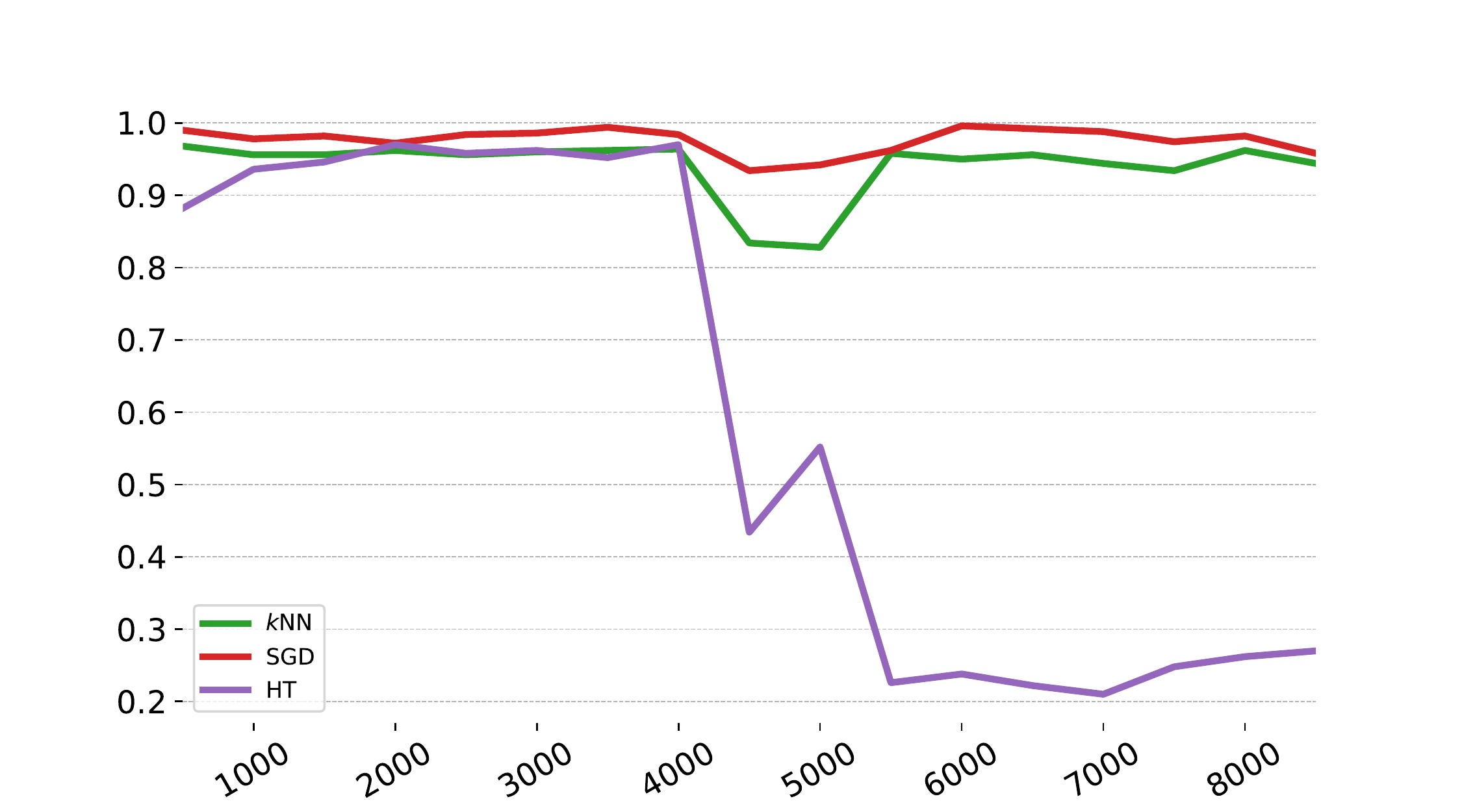}
	}
	\subfloat[][Gradual drift]{
		\label{f:synth.d}
		\includegraphics[width=0.44\textwidth]{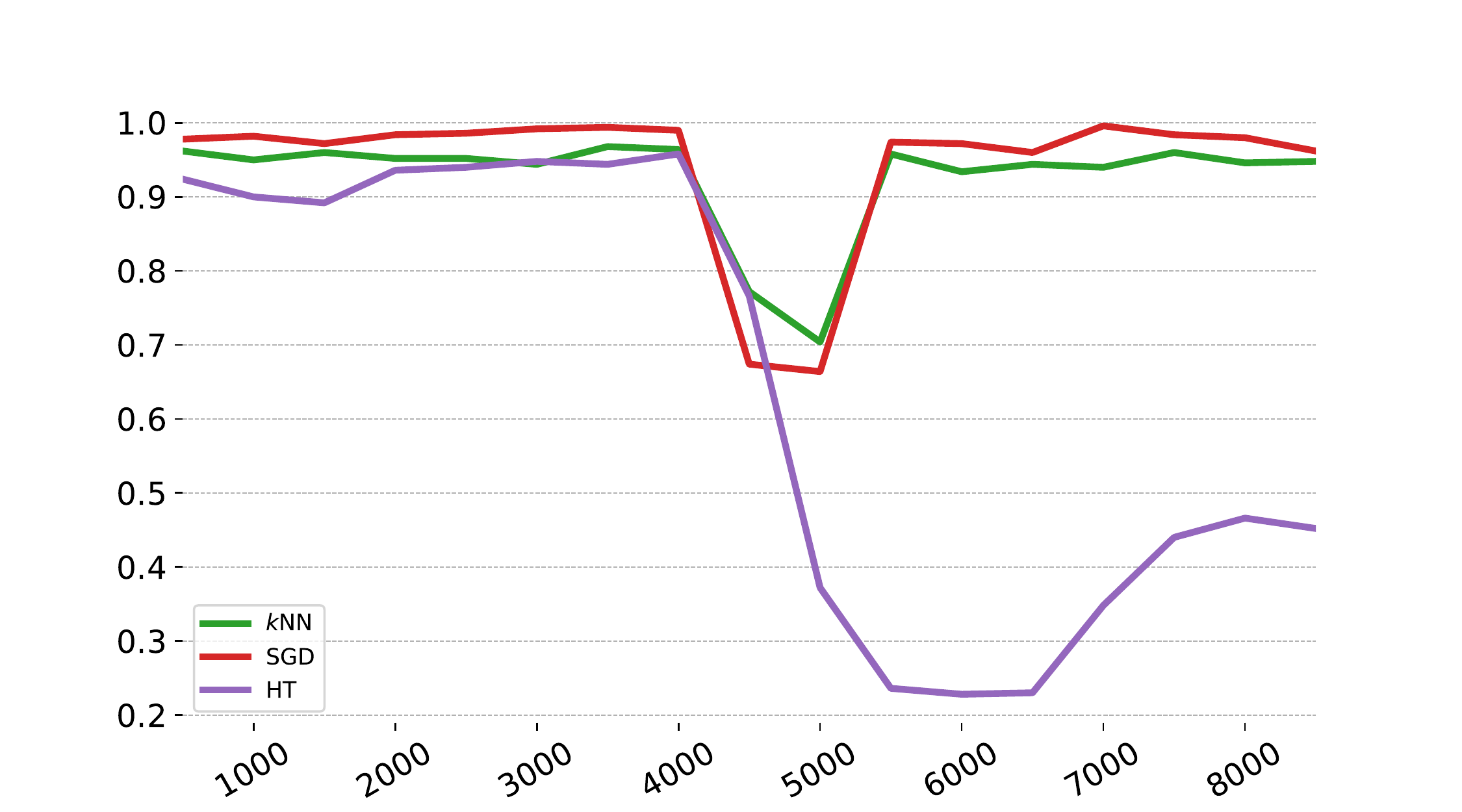}
	}
	\caption{\label{fig:synth}Results (classification accuracy over a sliding window of $200$ instances on synthetic streams. Notice that the vertical axes are scaled for greater visibility of separation.}
\end{figure*}

First we had a look at performance on the synthetic data (\Fig{fig:synth}), then on real world data (\Fig{fig:difficult}), and 
finally, on an especially challenging scenario (\Fig{fig:syncremental}) over a sustained incremental drift. 
\Tab{tab:accuracy} summarizes the predictive performance and \Tab{tab:time} provides some running time results.

\section{Discussion}
\label{sec:discussion}

We present and discuss results from the experiments outlined in the previous section. 



\subsubsection*{Hoeffding Trees are fast but conservative}

We observe how HTs grow conservatively (\Fig{f:synth.a}). This behaviour is intended by design, since they have no natural forgetting mechanism and thus it is important to make a statistically safe split (based on the Hoeffding bound). 

Indeed, note after $t=7000$ accuracy jumps 10 percentage points; indicating the initial split. 

This conservative approach provides strong confidence that we may produce tree equivalent to one built from a batch, but only within a single concept. It necessarily means that Hoeffding trees will struggle when the true target concept $\theta_t$ is a \emph{moving} target rather than a fixed point in concept space.

And we observe this: Figures \ref{f:synth.b}---\ref{f:synth.d} show how the performance of a standalone HT is damaged by concept drift (as opposed to SGD- and $k$NN-based methods which are able to recover more quickly. \cout{If a new split is made, recovery can be carried out to some extent, but unlikely to be a complete solution.}

\subsubsection*{Destructive adaptation is costly}

\begin{figure*}[h]
	\centering
	\includegraphics[width=1.0\textwidth]{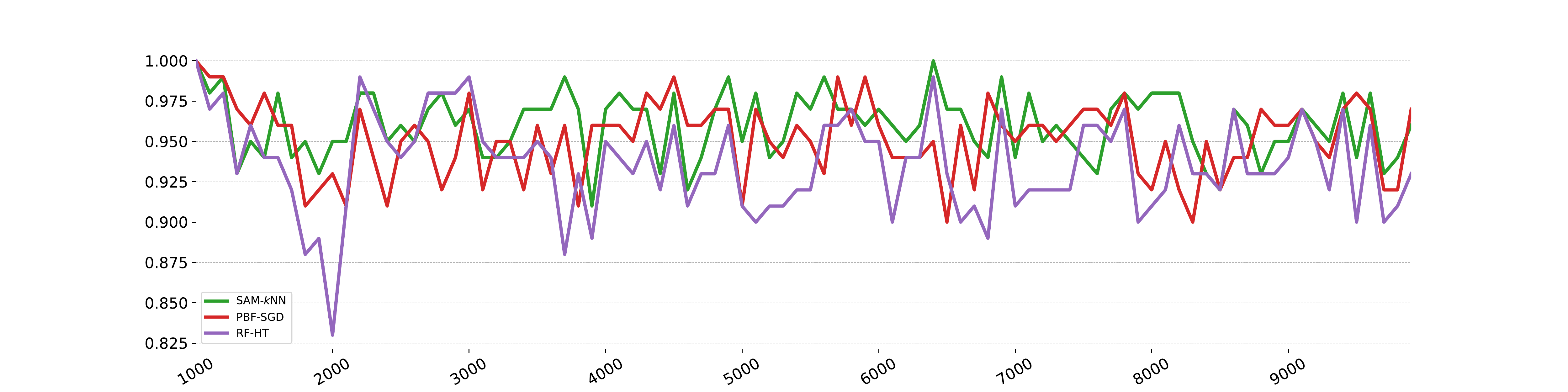}
	\caption{\label{fig:syncremental}An incremental drift scenario like that of \Fig{fig:synth} except that the drift is constant across the entire stream ($\tau_1=\tau_0$, and $\tau_2=T$ wrt \Tab{tab:synth}).}
\end{figure*}

\begin{figure*}[ht]
	\centering
	\subfloat[][Electricity]{
		\includegraphics[width=0.90\textwidth]{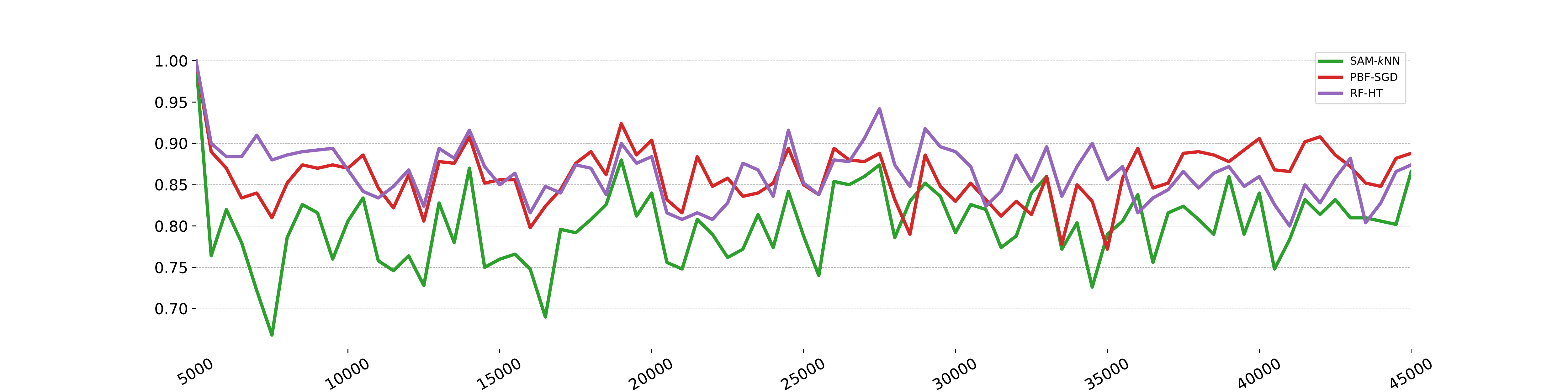}
	}\\
	\subfloat[][CoverType]{
		\includegraphics[width=0.90\textwidth]{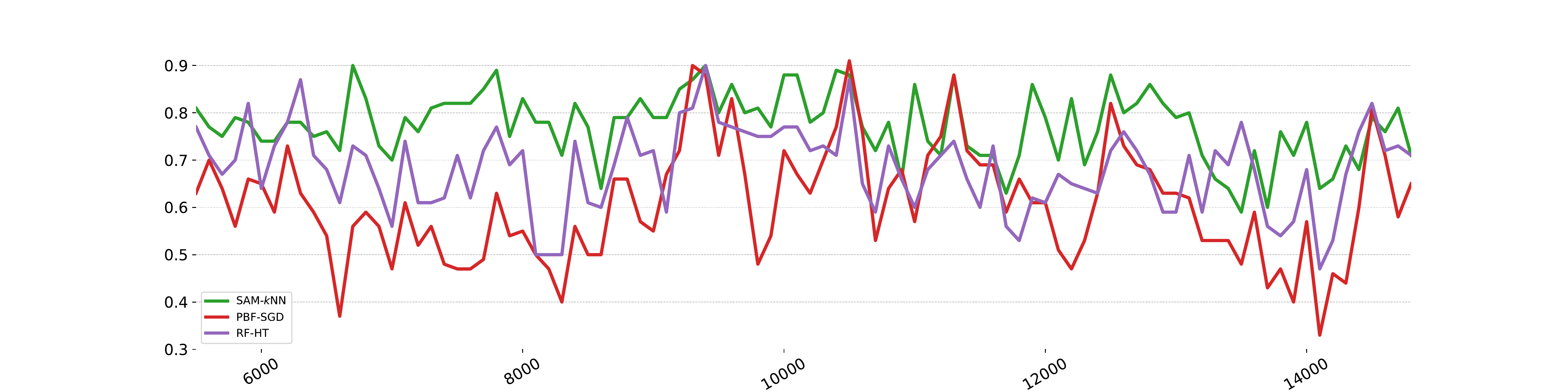}
	}\\
	\subfloat[][Random Tree Generated (RTG) stream]{
		\includegraphics[width=0.90\textwidth]{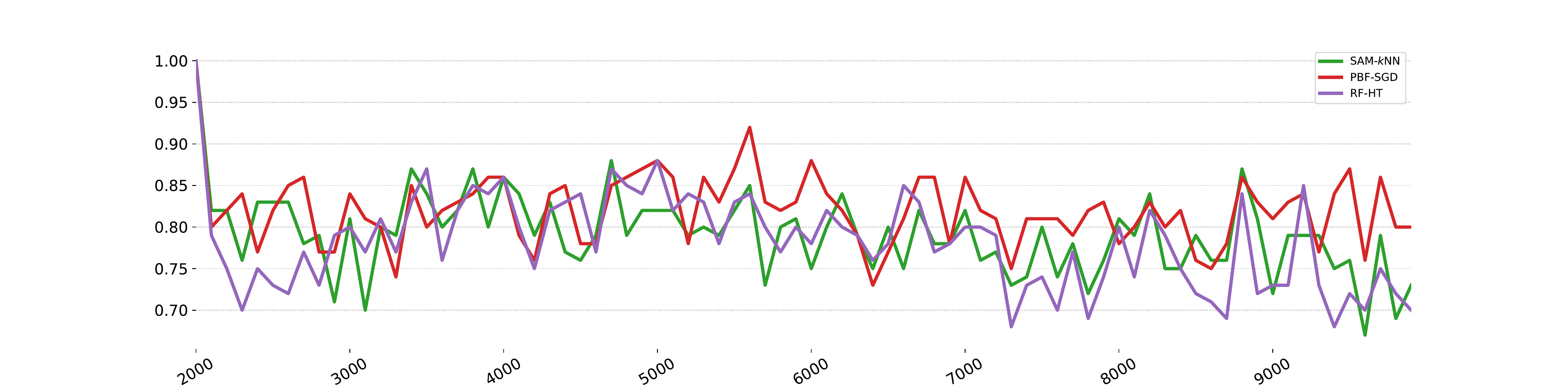}
	}
	\caption{\label{fig:difficult} Average accuracy over sliding window of 200 for methods detailed in column 3 of \Tab{tab:methods}. For visualization, we plot only a subset of CoverType. See \Tab{tab:accuracy} for total accuracy over the streams.}
\end{figure*}

In the literature sophisticated concept-drift detectors and ensembles are used to counteract this disadvantage, and provide robust prediction in dynamic data streams. And we confirm that this approach is effective in many cases (as seen in \Fig{fig:difficult}) 
but at a significant cost: 
\cout{Such approaches are not parsimonious, but rather overcome limitations by force of ensembles in a kind of creative destruction. We can see this this in \Fig{fig:syncremental}, where advanced methods (see \Tab{tab:methods} column 3 for respective parameterizations) including the adaptive forest of Hoeffding trees (RF-HT), are deployed on a stream of constant drift. Despite a state-of-the-art drift detection and ensemble adaptation scheme, this powerful ensemble is unable to achieve higher performance than a relatively simpler single-model SGD-based method. Whereas the ensemble involves ongoing destruction of decision trees in order to catch up with new dynamics of the stream, SGD adapts to the drift in an instance-by-instance fashion. Accuracy alone does not distinguish any of the three, methods, but the difference in}
the computational time (see \Tab{tab:time}) of the detect-and-reset approach clearly enunciates the overhead of destroying (resetting) and regrowing HTs constantly, so as to adapt to a changing concept. 




\cout{
	The synthetic data used in initial experiments is based on a rotating decision line, for illustration of several mechanisms. We also compare more powerful renditions of each approach on on real-world benchmark datasets and the RTG stream that is certainly expected to favor tree-based methods. Results are shown graphically in \Fig{fig:difficult} over a window, and tabulated in \Tab{tab:accuracy}.}

\cout{As specified above (\Sec{sec:proc}, \Tab{tab:methods})  we compare to the recent and state-of-the-art adaptive random forest (RF-HT) and self-adjusting memory $k$NN (SAM$k$NN); and we employ a relatively simple polynomial basis-function expansion for SGD (calling it thus PBF-SGD).} 

\subsubsection*{Buffer-based methods are limited}

Respective of its buffer size, $k$NN-methods can respond to drift by forgetting old instances (from old concepts) and taking in (i.e., learning from) new ones. 
This mechanism allows it to recover quickly from drift (\Fig{f:synth.b}--\Fig{f:synth.d}). Nevertheless predictive power is always limited in proportion to the number of instances stored in this buffer and, as observable in \Subfig{f:synth.a}, sometimes this is insufficient (we note there is no upward trend here, as opposed to the other methods, despite more instances from the same concept). If buffer size is widened, performance can be higher, but adaptation to drift will take correspondingly longer or require explicit drift detection methods as with HT approaches. 


\subsubsection*{SGD for Efficient Continuous Adaptation}

SGD is a simple method which has been around a long time. With a non-decayed learning rate, we find that it behaves as well as we hypothesized on synthetic data: it continues to learn a static concept better over time (in \Fig{f:synth.a} it recovers quickly from sudden and gradual drift, and its performance is almost unaffected under incremental drift where (as we see in \Fig{f:synth.c}) it is able to adapt continuously.

We suspect that SGD has not been widely considered in state-of-the-art data-stream evaluations because it performs poorly on real-world and complex data when deployed in an off-the-shelf manner, especially if the learning rate is decayed -- as is often the standard. However, we put together the PBF-SGD method from elementary components and find that it performs strongly in these scenarios (\Fig{fig:syncremental}, \Fig{fig:difficult}). 

We do see that performance of the advanced/state-of-the-art methods (HT and $k$-NN based) is also competitive, as expected, yet it is crucial to emphasise the difference in computational performance (see \Tab{tab:time}): running time is up to an order of magnitude or more higher for the decision tree ensemble, compared to other methods; even greater than PBF-SGD, which has a feature space cubicly greater than the original. 

\subsubsection*{An analysis of time and memory complexity}

The worst-case complexity is outlined in \Tab{tab:complexity} (for the vanilla methods, which does not take into account the additional overhead of ensembles and drift detectors nor basis expansion for PBF-SGD). 
We can further remark computational time and memory expectations of SGD are constant across time (the expected running time is the same for each instance in the stream), as also with $k$NN (given a fixed batch size). On the other hand, HT costs are not constant, but continue to grow with the depth of the tree (the $\ell$ term in \Tab{tab:complexity}). This is an issue which as, to our knowledge, not been considered in depth in the literature: as trees in an ensemble grow and are reset under drift, time and space complexity fluctuates -- making practical requirements difficult to estimate precisely in advance. If there is no drift -- then the trees may, in theory, grow unbounded and use up all available memory. 

	\begin{table}
		\centering
		\caption{\label{tab:accuracy}Overall accuracy (average over entire stream) of methods; seen also in experiments of \Fig{fig:syncremental} (regarding the Synthetic stream) and \Fig{fig:difficult} for the others.}
			\begin{tabular}{llll}
				\hline
								         & SAM$k$NN & PBF-SGD       & RF-HT  \\
				\hline
				Electricity              & 79.8    & 85.9         & 86.2  \\  
				RTG                      & 78.8    & 81.8         & 77.9  \\
				CoverType                & 93.3    & 92.6         & 93.9 \\
				Synthetic & 96.0    & 95.1         & 93.6 \\
				\hline
			\end{tabular}

	\end{table}

	\begin{table}[h]
		\centering
		\caption{\label{tab:complexity}An outline of time and space complexity for different methods, with a moving window of $w$ examples, taking the $k$ nearest neighbors (for $k$NN), in a problem of $d$ attributes. For HT, $\ell$ is the number of leaves, but note that this is $O(n)$ in the worst case. To simplify, we have considered a binary ($2$ class) problem with one split considered per attribute in the case of trees.}
		\begin{tabular}{lll}
			\hline
			Method & Time     & Space \\
			\hline
			$k$NN    & $O(wdk)$ & $O(wd)$ \\
			HT     & $O(d)$  & $O(\ell d)$ \\
			SGD    & $O(d)$   & $O(d)$ \\
			\hline
		\end{tabular}
	\end{table}

\cout{Furthermore, we add that there is no theoretical limitation to gradient descent methods in terms of concept modelling -- even one additional feature layer of non-linear functions can be a universal approximator \cite{UniversalApproxRBF}. Therefore, with the right basis functions in sufficient quantity it is possible to represent any concept.}

\begin{table}[ht]
	\centering
	\includegraphics[width=0.5\columnwidth]{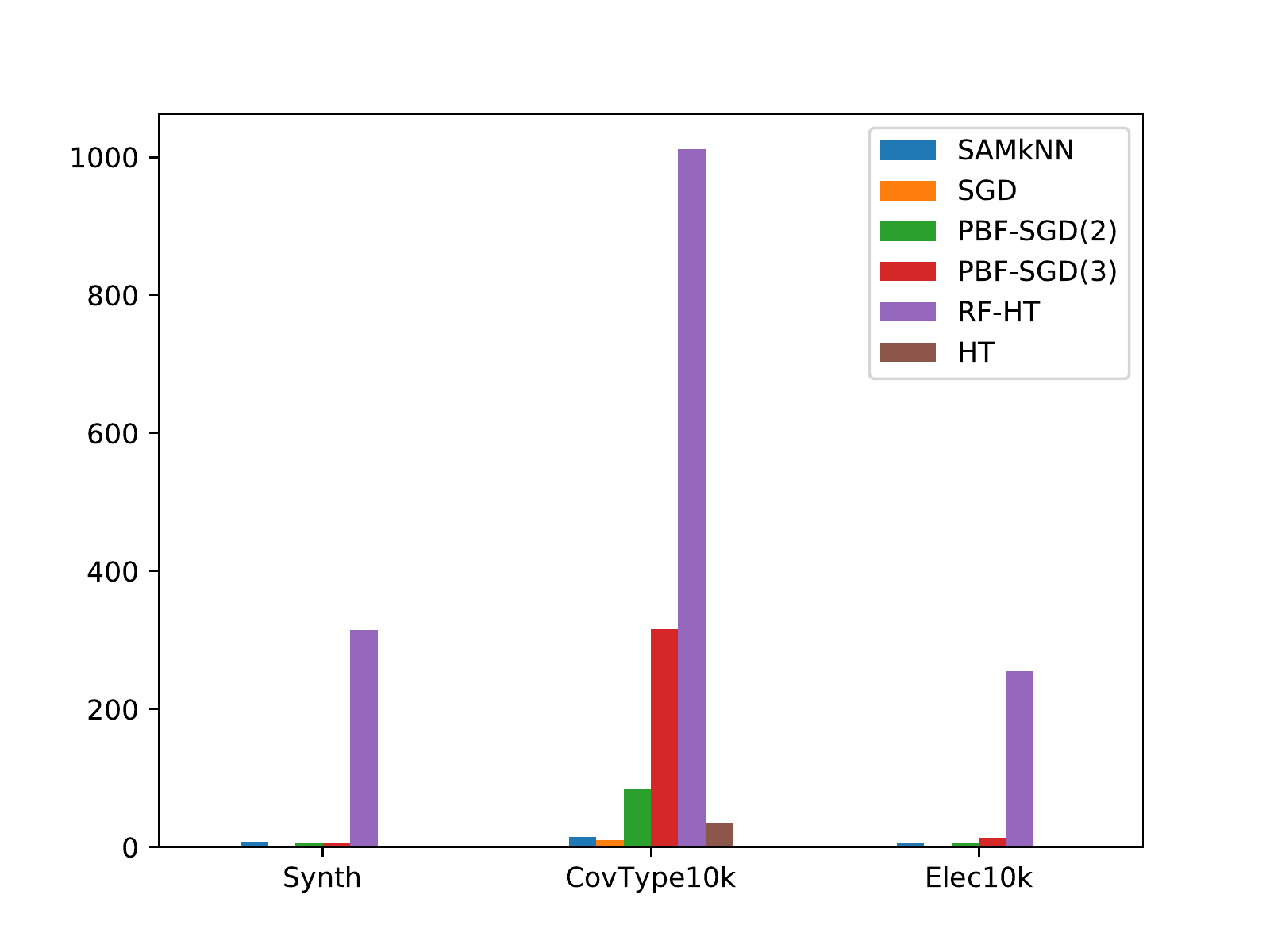}
	\caption{\label{tab:time}Total running time under prequential evaluation (in seconds) on a selection of the datasets. We compare SGD with both polynomial of degree 2 and 3 (PBF-SGD(2) and (3), respectively) -- for sake of comparison. \textsf{Synth} refers to the data of \Fig{fig:syncremental}. The time including the initial block (which is not evaluated in terms of accuracy). Note that for all timing experiments we only considered the first 10,000 instances.}
\end{table}








\subsubsection*{Limitations and Further Considerations}

We need to acknowledge that an incremental decision-tree based approach continues to be a powerful competitor, and more efficient implementations exist than the Python framework we used in this work. Furthermore, it is clear that more work is needed to investigate performance under sudden and gradual and mixed types of drift. There are other state-of-the-art HT methods as well as $k$NN methods which could be additionally experimented with. 

However, we have shown analytically and empirically that the most desirable and efficient option is supported naturally by gradient descent (and by extension, neural networks), given certain constraints wrt the learning rate, namely that is significantly greater than zero. This indicates that more attention in the streaming literature should be paid to neural networks, in particularly on ways to parameterize them effectively for data-stream scenarios, so that they are more easy to deploy.  

In general it is a more promising strategy to model the drift and pre-empt its development, rather than waiting for an indication that drift has already occurred and following such an indication retrospectively, to reset models that have been previously built. 


%
%




\section{Conclusions} 
\label{sec:conclusion}

The literature for data streams is well furnished with a great number and diversity of ensembles and drift-detection methodologies for employing incremental decision tree learners such as Hoeffding trees. These methods continue to obtain success in empirical evaluations. In this paper we have taken a closer analytical look at the reasons for this performance, but also we have been able to highlight its limitations, namely the cost involved of its performance under sustained drift, where it forced to carry out a continued destructive adaptation. 

In particular, we showed that concept-drifting data streams can be treated as time series, which in turn suggests predictability, thus encouraging an approach of tracking and estimating the evolution of a concept, and carrying out continuous adaptation to concept drift. To demonstrate this we derived an appropriate approach based on stochastic gradient descent. The method we used was simple, but results clearly supported our analytical and theoretical argument which carries important implications: gradient-based methods offer and effective and parsimonious way to deal with dynamic concept drifting data streams and should be considered more seriously in future research in this area. This is especially true with the advent of more powerful and easy-to-deploy neural networks and recent improvements in gradient descent. 

In the future will investigate more complex scenarios involving oscillating and mixtures of different times of drift, and experiment with more state-of-the-art gradient-descent approaches, such as deep learning. Our study already indicates that further investigation along these lines will yield promising results.






\bibliographystyle{plain}
\bibliography{my_publications,datastreams,multilabel,other}
%

\end{document}